\documentclass[letterpaper, 10 pt, conference]{ieeeconf}  

\IEEEoverridecommandlockouts                              

\usepackage{times}
\usepackage{epsfig}
\usepackage{graphicx}
\usepackage{amsmath}
\usepackage{amssymb}
\usepackage{xspace}

\usepackage{amsfonts}

\usepackage{epstopdf}
\usepackage{color}

\usepackage{multirow}
\usepackage{pbox}

\usepackage{amsopn}

\title{\LARGE \bf
Track, then Decide: Category-Agnostic Vision-based \\Multi-Object Tracking
}

\author{Aljo\v{s}a O\v{s}ep, Wolfgang Mehner, Paul Voigtlaender, and Bastian Leibe
\thanks{\scriptsize The authors are with the Visual Computing Institute, RWTH Aachen University
\newline
E-mail: {\tt\scriptsize lastname@vision.rwth-aachen.de}}%
}

\makeatletter
\DeclareRobustCommand\onedot{\futurelet\@let@token\@onedot}
\def\@onedot{\ifx\@let@token.\else.\null\fi\xspace}

\def\eg{\emph{e.g}\onedot} 
\def\ie{\emph{i.e}\onedot} 
\def\cf{\emph{c.f}\onedot} 
\def\etc{\emph{etc}\onedot} 
 
\def\etal{\emph{et al}\onedot}

\newcommand{\PAR}[1]{\vskip4pt \noindent {\bf #1~}}

\DeclareMathOperator*{\argmax}{\arg\!\max}
\DeclareMathOperator{\iou}{\text{IoU}}
\DeclareMathOperator{\pose}{\mathbf{p}}
\DeclareMathOperator{\velocity}{\mathbf{v}}

\DeclareMathOperator{\bbox_2d}{\text{box}}
\DeclareMathOperator{\mask}{\mathbf{m}}

\newcommand{\hypo}[2]{\ensuremath{\operatorname{\emph{h}^{#1}_{#2}}}}

\newcommand{\hyposet}{\ensuremath{\operatorname{\mathbf{h}}}}
\newcommand{\observset}[2]{\ensuremath{\operatorname{\mathbf{O}^{#1}_{#2}}}}
\newcommand{\obs}[2]{\ensuremath{\operatorname{o^{#1}_{#2}}}}

\newcommand{\binvec}{\ensuremath{\mathbf{b}}}

\begin{document}

\maketitle
\thispagestyle{empty}
\pagestyle{empty}

\begin{abstract}

The most common paradigm for vision-based multi-object tracking is tracking-by-detection, due to the availability of reliable detectors for several important object categories such as cars and pedestrians.
However, future mobile systems will need a capability to cope with rich human-made environments,
in which obtaining detectors for every possible object category would be infeasible.
In this paper, we propose a model-free multi-object tracking approach that uses a category-agnostic image segmentation method to track objects.
We present an efficient segmentation mask-based tracker which associates pixel-precise masks reported by the segmentation.
Our approach can utilize semantic information whenever it is available for classifying objects at the track level, while retaining the capability to track generic unknown objects in the absence of such information.
We demonstrate experimentally that our approach achieves performance comparable to state-of-the-art tracking-by-detection methods for popular object categories such as cars and pedestrians.
Additionally, we show that the proposed method can discover and robustly track a large variety of other objects.

\end{abstract}

\section{INTRODUCTION}

Multi-object tracking (MOT) is one of the most critical components in visual scene understanding for mobile platforms, such as autonomous cars and robots. If such systems are to perform safe navigation and motion planning in urban environments, they need to be aware of surrounding objects and be able to predict their future motion.
\begin{figure}
\centering{
	\includegraphics[width=0.5\textwidth]{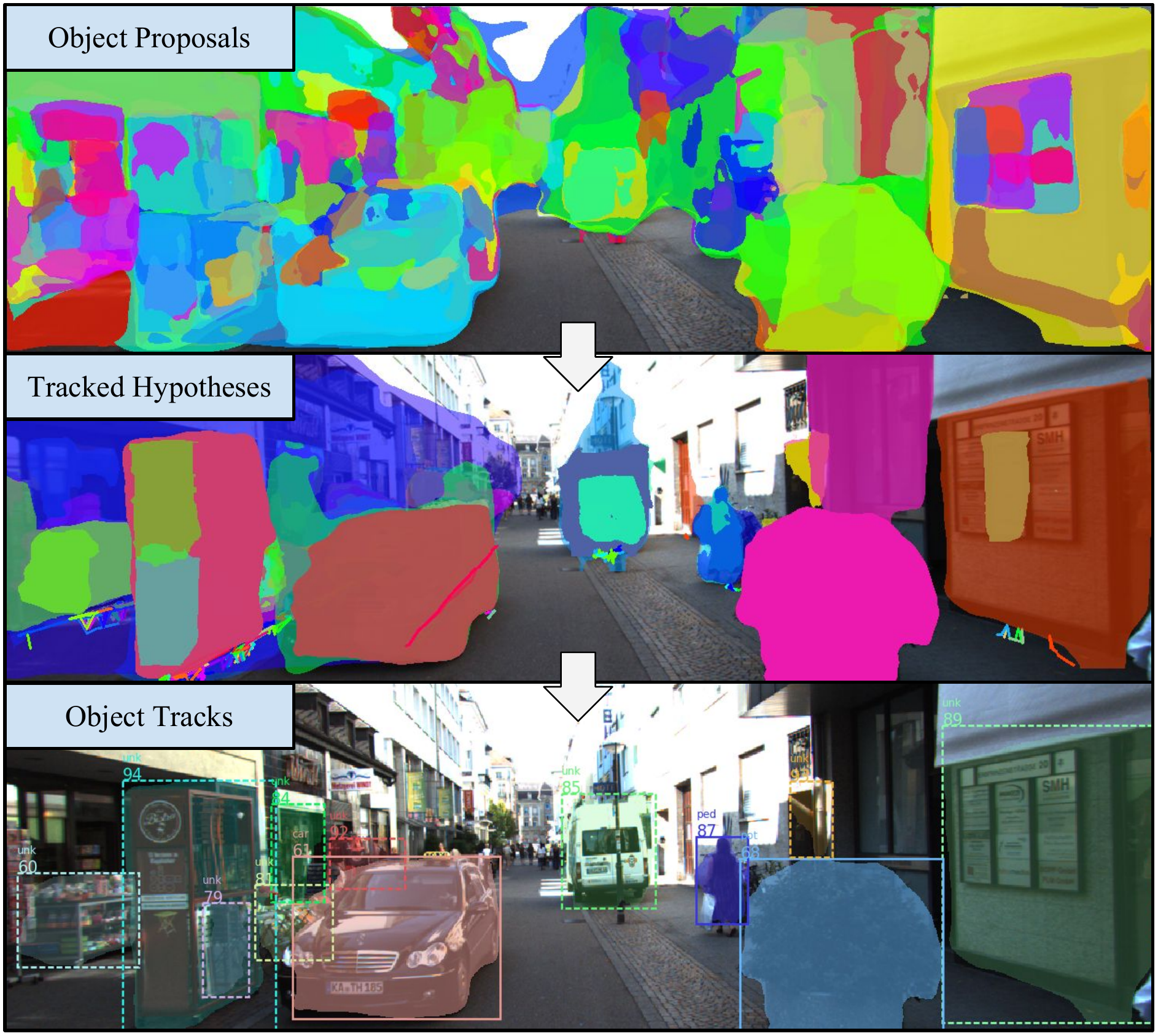}}
	\caption{Our proposed method can handle a large number and variety of generic object proposals simultaneously.
	It robustly tracks objects of both known and unknown categories in challenging street scenarios and reports a manageable set of relevant objects.}
	\label{fig:screenshot_seq10}
\end{figure}

In recent years, notable successes have been achieved for vision-based tracking of the most common traffic participant classes, such as cars and pedestrians. This development has largely been driven by improvements in object detectors \cite{RenNIPS15,Redmon16CVPR,Liu16ECCV} and tracking-by-detection approaches, \eg \cite{Choi15ICCV,Geiger12CVPR,Osep17ICRA,Wang15BMVC,Xiang15ICCV,Yoon16CVPR}.
However, there is another challenge that needs to be addressed before mobile systems using vision sensors can safely operate in human-made environments:
these environments are full of a large variety of other objects for which reliable detectors may not easily be obtained.

In LiDAR-based multi-object tracking for autonomous driving,
\eg~\cite{Teichman2012IJRR, Moosmann13ICRA}, this problem is addressed by object-instance segmentation of LiDAR scans.
These instances are then associated from frame to frame and are classified on a trajectory level.
Using such approaches, a larger variety of objects can be tracked.
However, current LiDAR sensors suffer from a very limited resolution, making recognition difficult. Stereo vision can provide a far richer input signal; however, reliably extracting candidate objects from noisy stereo data is a much harder problem.

In this paper, we address class-agnostic multi-object tracking using only an inexpensive stereo setup.
Our proposed approach starts by extracting a large number of object region proposals from the input images and tracks them over time using both 2D and 3D information.
The basic assumption behind our method is that correct object region hypotheses will lead to more consistent tracks and thus to a better explanation for the observed scene change.
Thus, we use tracking consistency as a cue to identify ``interesting'' objects.
Our approach provides as output a hypothesized trajectory for each such object.
We then apply a region classifier (in our case the one from Faster R-CNN \cite{RenNIPS15}) to each hypothesized track in order to provide a semantic label for known object categories, which also helps resolve ambiguities between overlapping object hypotheses.
This allows our approach to achieve competitive tracking performance for the most common categories of traffic participants, while retaining the capability to track unknown objects.

The main challenge behind such a generic multi-object tracking approach is the large number of object region proposals and the variety of shapes it needs to consider, since it cannot rely on an initial object detector.
To not introduce further problems by wrong data associations, we require a very precise, pixel-level region tracking procedure as its main ingredient.

In this paper, we propose a region tracking method that relies on pixel-precise masks to perform the frame-to-frame alignment. The masks are stored and compared efficiently using a compressed representation with run-length encoding (RLE).
In addition, we track the 3D position of each potential object, obtained from the stereo point cloud, for precise 3D localization.
Finally, our approach takes into account category predictions from a region-based object classification method, wherever available, in order to help resolve ambiguities between overlapping objects.
Altogether, this allows our approach to robustly track a large number of known and objects in challenging street scenes.

Our main contributions are:
(i) We present a vision-based tracking approach that performs segmentation-based, category-agnostic multi-object tracking.
(ii) The core component of our approach is a novel and efficient segmentation mask-based region tracker that utilizes a coarse geometric scene understanding to predict the shape and position of the masks in future frames, and which uses this information for pixel-precise data association.
(iii) Our approach can benefit from additional semantic information for known object classes from a region classifier which we apply to each object hypothesis.
Applying a classifier after tracking instead of tracking detections enables us to achieve the capability of model-free tracking when no semantic information is available.
We show experimentally that our method is competitive with state-of-the-art tracking-by-detection methods for cars and pedestrians at a distance range of up to 30m. In addition, we demonstrate that it generalizes well to a large variety of other object categories.

\section{Related Work}

\PAR{Vision-based Multi-Object Tracking in Street Scenes.} The typical vision paradigm for multi-object tracking in street scenes is tracking-by-detection \cite{Zhang08CVPR, Milan14TPAMI, Leibe08TPAMI}. Current state-of-the-art methods focus on learning optimal parameters for data association \cite{Wang15BMVC, Xiang15ICCV}, designing powerful affinity measures \cite{Choi15ICCV} or explicitly dealing with abrupt camera motion \cite{Yoon16CVPR, Osep17ICRA}. Apart from \cite{Osep17ICRA}, all these methods perform tracking only in the image-domain and are therefore not directly applicable to robotics applications.

\PAR{Model-free Object Tracking.} In contrast to the typical vision tracking paradigm, in robotics a reverse pipeline is often applied: tracking-before-detection, where the tracking process is category-agnostic. Motion detection-based methods such as \cite{Kaestner12ICRA, Moosmann13ICRA, Dewan15ICRA} detect and track moving objects in LiDAR data.
The approaches by \cite{Teichman2012IJRR, Teichman13ICRA, Teichman11ICRA} use a pipeline based on LiDAR point cloud segmentation. First, a coarse preprocessing is applied, including ground/building facade removal, followed by a grouping of the remaining points that are then fed to the tracker. The resulting trajectories are then verified by a classifier using features extracted from the trajectory and LiDAR segments. An important property of such approaches is that they have the capability of tracking objects, even when their categories cannot be recognized, since LiDAR measurements are a strong evidence of object presence. However, with such LiDAR clustering-based methods, under- and over-segmentations still occur due to occlusions, reflective and low-albedo surfaces and sparsity increasing with distance. To address these issues, Held \etal \cite{Held16RSS} propose a method that uses spatial, temporal and semantic cues to reduce the number of over- and under-segmentations.

There have been a few attempts to employ similar ideas in the vision community. Mitzel and Leibe \cite{Mitzel12ECCV} propose an approach for tracking pedestrian-sized objects based on stereo point cloud segmentation using Quick-Shift \cite{Vedaldi08ECCV} and an ICP tracker \cite{Besl92PAMI}.
Lenz \etal~\cite{Lenz2011IV} propose a method for model-free multi-object tracking of moving objects based on sparse scene flow clustering.
\cite{Zhu16ACCV} and \cite{Osep16ICRA} utilize several, possibly overlapping object proposals to track generic objects. \cite{Osep16ICRA} propose a two-stage segmentation using stereo data. First, a coarse removal of ``background'' categories is performed; second, the remaining ``object'' points are clustered using a multi-scale variant of \cite{Mitzel12ECCV} and are fed into a tracker. Zhu \etal~\cite{Zhu16ACCV} use object proposals to learn appearance models online for model-free multi-object tracking.

\PAR{Object Proposal Generation.}
In the context of LiDAR-based object instance segmentation, Teichman \etal~\cite{Teichman2012IJRR} project point-clouds on the estimated ground-plane and clusters the points using a simple flood-fill algorithm. Wang \etal~\cite{Wang12ICRA} propose to perform first a coarse classification of points into foreground and background regions and cluster remaining foreground regions using graph-based clustering (similar to \cite{Osep16ICRA}). Bogoslavskyi and Stachniss \cite{Bogoslavskyi16IROS} propose a very efficient LiDAR segmentation method that operates directly on a 2D range image. \cite{Kochanov2016IROS} generates object proposals by integrating stereo point clouds and coarse semantic information over time in a 3D voxel grid. Cells corresponding to the ``object`` class are then clustered using DBSCAN \cite{Ester96KDD}.

Traditional image-based methods for object proposal generation and instance segmentation rely on grouping low-level segmentation cues (\eg superpixels) and hand-crafted features \cite{Alexe12TPAMI, Zitnick14ECCV, Uijlings2013IJCV}.
Recent methods leverage convolutional neural networks (CNNs) trained on large datasets with mask-based annotations (\eg COCO \cite{Lin14ECCV}).
Pinhero \etal~\cite{Pinhero16NIPS, PinheiroECCV16} propose an object proposal generation method using CNNs, learned end-to-end. Their SharpMask method \cite{PinheiroECCV16} samples rectangular patches at multiple scales. For each patch, the network outputs a  class-agnostic segmentation mask and a probability estimate of how likely the center of the patch contains an object. By training their network on several subsets of the object categories, they demonstrate that SharpMask generalizes to unseen categories and is thus category-agnostic.

\cite{Horbert2015ICRA} proposes a method for generating spatio-temporal object proposals in videos by tracking several, possibly overlapping per-frame object proposals using level sets. Short, stable tracklets then form object candidates. Most recent methods propose to learn to generate such sequence-level-based proposals (bounding-box-based) in an end-to-end fashion \cite{Kang17CVPR}. While abovementioned are similar to our method, their ultimate goal is to produce ``smoother`` object proposals in video sequences. Multi-object tracking (MOT) is a higher-level task. MOT methods need to maintain object identities, gap occlusions, and, especially for robotics applications, additional information like the 3D pose, velocity, and object size estimates need to be provided.

\section{Our Approach}

\begin{figure*}[t!]
\begin{center}
\includegraphics[width=0.9\linewidth]{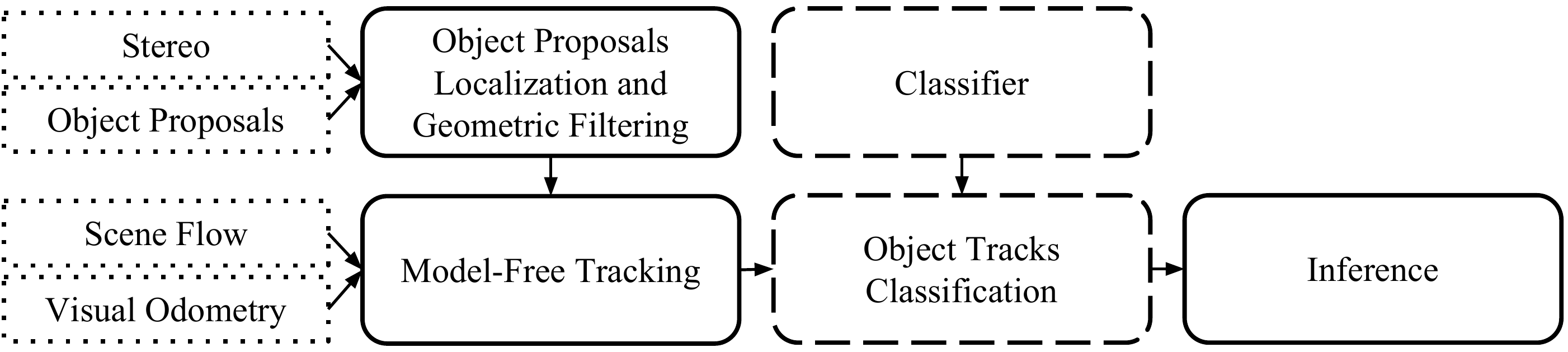}
\end{center}
\caption{Our tracking method receives a large set of region proposals, creates a set of ``proposal'' object hypotheses via tracking,
	and finally selects those that are the most likely explanation of the observed cues.
	Optionally, a classifier can be used to identify the object category.
	}
\label{fig:pipeline}
\end{figure*}

Fig.~\ref{fig:pipeline} shows an outline of our approach.
Instead of using an object detector we obtain proposal regions,
which are represented as pixel-precise segmentation masks, one for each proposed object instance.
We require that enough of these masks correspond to actual objects, and that the reported regions are reasonably stable across frames.
While our approach can handle such inputs from different sources, we use SharpMask \cite{PinheiroECCV16} here.
The region proposals are then located in 3D using depth from stereo.
These observations are filtered with our novel integrated 2D-3D tracking approach.
It is able to cope with the large number of tracking hypotheses, resulting from the large number of class-agnostic proposal regions.

We maintain an over-complete set of competing spatio-temporal tracking hypotheses.
Using inference in a Conditional Random Field (CRF) model following the \textit{hypothesize-and-select} paradigm \cite{Leibe08TPAMI, Choi15ICCV, Osep17ICRA}, we choose valid hypotheses for reporting.
Additionally, our proposed model can make use of semantic information in the form of class labels during inference.
Knowledge of semantic information helps to resolve segmentation ambiguities, but our approach maintains the capability to keep tracking objects that are not recognized by the classifier.

The SharpMask region proposals help with our goal of class-agnostic tracking, by providing consistent segmentation masks.
But it is only by applying our proposed tracking pipeline, and utilizing a combination of 2D and 3D cues, that we are able to cope with the huge number of over- and under-segmentations included in the region proposals.
In the inference step we can then select a meaningful subset of these hypotheses for reporting,
by turning the consistent region proposals into consistent tracking hypotheses, and rejecting inconsistent proposals which do not result in strong hypotheses.

%
%

\subsection{Segmentation Mask Localization in 3D Space}

We begin with the set of region proposals from SharpMask \cite{PinheiroECCV16}, which is then reduced by applying non-maximum suppression.
We first sort their masks by their score, and then suppress them using intersection-over-union as a similarity measure.
Afterwards, we perform a coarse geometric filtering of the segmentation masks.
To this end, we fit a ground plane to the stereo point cloud using RANSAC \cite{Fischler81ACM}.
Then we compute the median 3D position and velocity of the segmentation masks using stereo and scene flow estimates by \cite{Vogel13ICCV}.
We filter out object proposals that are below and above certain height thresholds, keeping only the objects that ``stick out'' of the ground (with some tolerance, as objects may be occluded).
The remaining masks, together with position and velocity estimates, represent the inputs (observations) to our tracker.

In each frame $t$, we obtain a set of observations $\observset{t}{} = \left \{ \obs{t}{k}  | k \in \left \{ 1 \ldots K \right \} \right \}$.
Each observation is defined by $\obs{t}{k} = \left [ \pose, \velocity, \mask, s \right ]$, where $\pose = \left [ x, y, z \right ]^T$ represents its 3D position, $\velocity = \left [ \dot{x}, \dot{z} \right ]^T$ represents its velocity (only $x$ and $z$ components) and $\mask$ represents the segmentation mask.
Finally, $s$ represents the mask score (how likely the mask represents an actual object, this is an additional output of SharpMask).
In practice, we limit $K$ to $100$.

\subsection{Segmentation-based Multi-Object Tracking}
In our method we use the \textit{hypothesize-and-select} tracking paradigm \cite{Leibe08TPAMI},
which allows us to accumulate information over time.
We make delayed decisions as to whether a tracked object should be reported, based on the history of the hypotheses, rather than information from a single frame only.
The per-frame input to our method is a large set of segmentation masks, which contains correct segmentations, as well as over- and under-segmentations of the objects and false positives due to the background.
In order to cope with such an input, our tracker needs to be both efficient and precise, to avoid producing many false hypotheses due to segmentation clutter.
In this section, we present a simple, yet effective solution.

Our tracker maintains a set of overlapping hypotheses $\hyposet^t = \left \{ \hypo{t}{i} | \ i \in \left \{ 1 \ldots M \right \} \right \}$ by performing two operations in each frame:
(1) We use $\observset{t}{}$ to extend the existing hypotheses, see \textit{Data Association and Correction}.
(2) We start new hypotheses from each observation in the current frame in order to obtain a rich hypothesis set.
To initialize hypotheses, we perform data association and filtering backwards in time,
trying to equip a current observation with enough of a history to improve the estimation of the initial state.

\begin{figure}
\centering{
	\includegraphics[width=0.4\textwidth]{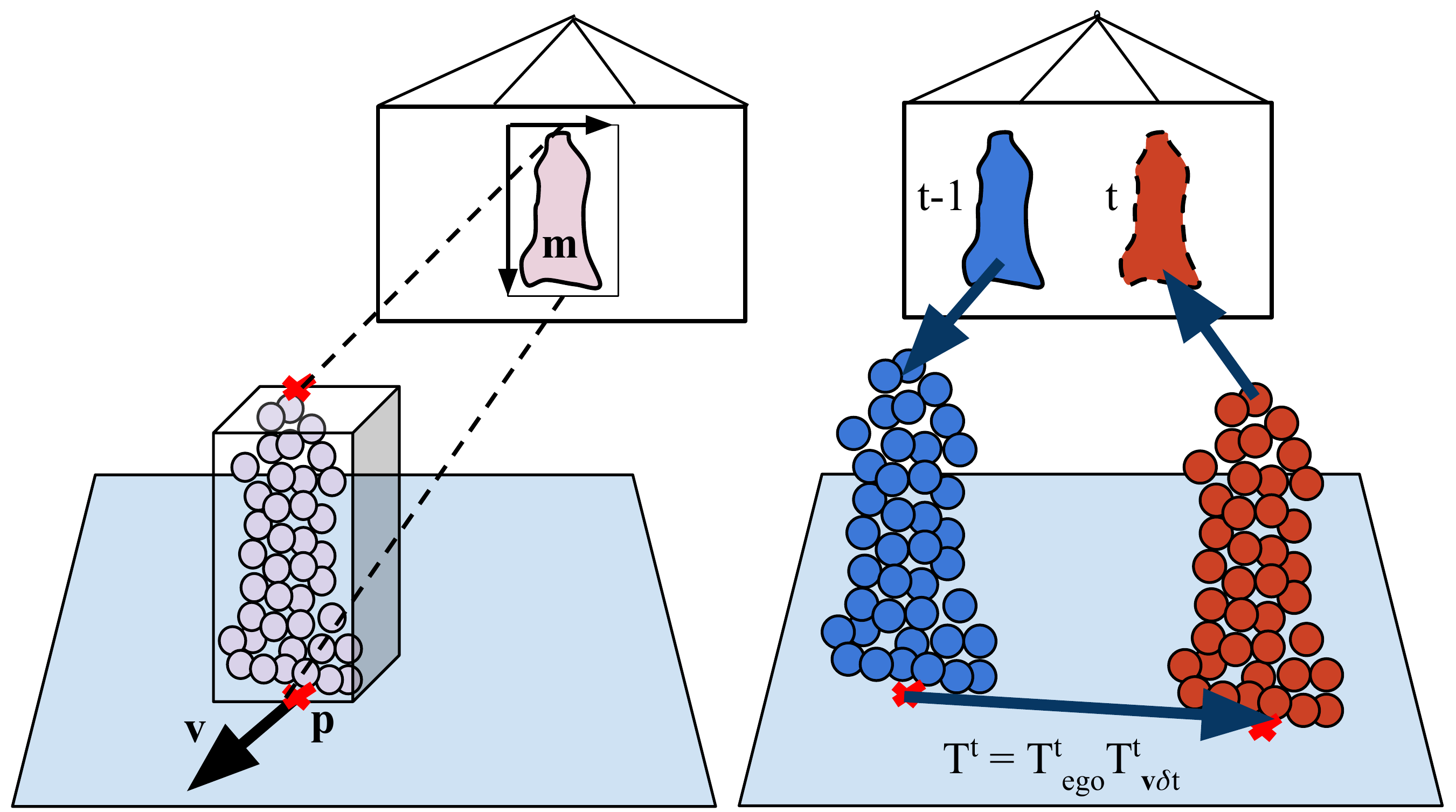}
	}
	\caption{(left) Hypothesis representation.
	Each hypothesis is represented by 3D position $\pose$ and velocity $\velocity$ vectors in 3D space and pixel-mask $\mask$ in the image.
	Bounding boxes can be trivially derived from the pixel mask in the image domain, and 3D space given the point cloud.
	(right) Mask prediction visualization.
	By leveraging ego-motion and velocity information (given in 3D space), we obtain pixel-precise appearance predictions of the tracked objects in the future frames.}
	\label{fig:mask_tracking}
\end{figure}

The hypotheses are represented by the following state:
\begin{equation}
\hypo{t}{i} = \left [ \pose_i^t, \velocity_i^t, \mask_i^t \right ],
\end{equation}
where $\pose^t_{i} = \left [ x, y, z \right ]^T$ and $\velocity^t_{i} = \left [ \dot{x}, \dot{z} \right ]^T$ represent the filtered 3D position and velocity of a hypothesis in world space,
and $\mask^t_{i}$ represents the mask of the object in the image plane, see Fig. \ref{fig:mask_tracking} (left).
This state is estimated using the set of observations associated with the hypothesis.

Note that often bounding boxes are used to represent the tracked objects (2D bounding boxes in image-based tracking or 3D bounding boxes in LiDAR/RGBD tracking).
Our mask-based representations are trivially convertible to 2D bounding boxes and, given depth data, to 3D bounding boxes.
\PAR{Prediction.} Prediction is performed in two steps.
First, the position and velocity is estimated by a Kalman filter.
The state at time $t$ is computed by applying a constant-velocity motion model to the state
$ \left[ \pose^t_{prio}, \velocity^t_{prio} \right] = \left[ \pose^{t-1}_{post} + \velocity^{t-1}_{post}, \velocity^{t-1}_{post} \right] $
(see Fig. \ref{fig:mask_tracking}). The mask is predicted using $\velocity^t_{prio}$ as follows:
\begin{equation}
\mask^{t}_{prio} = P \cdot T^t \cdot \mathcal{S}^{t-1} \left ( \mask^{*} \right ),
\end{equation}
where $\mask^{*}$ is either the last associated mask $\mask^{t-1}$ or the extrapolated mask $ \mask^{t-1}_{prio} $ in case there was no observation.
The operation $\mathcal{S}^{t-1} \! \left( \cdot \right)$ lifts the mask $\mask^{*}$ into the 3D point cloud of the last time frame.
The rigid-body transformation $T^t = T_{ego}^t \cdot T_{\velocity\delta t}^t$ corrects for the relative motion.
The ego-motion transform $T_{ego}^t$ is estimated via the visual odometry approach by \cite{Geiger11IV}.
The object's velocity $\velocity^t_{prio}$ is applied via $T_{\velocity}^t$, where the vertical component is set to zero, effectively constraining the velocity to translations on the ground plane.
The intrinsic camera matrix P performs the projection of the 3D points back into image space.

Note that temporal mask propagation and online 3D model estimation as in \cite{Held14RSS, Mitzel12ECCV, Osep16ICRA} could be beneficial.
However, solely using estimated velocities would result in a drift in the model estimate,
while doing more precise registrations of the segments would be computationally expensive.
With our approach, we only apply two rigid transformations in 3D space and back-project the points to the image.
This is computationally efficient and provides us with precise appearance predictions in the image domain.

%
%
\PAR{Data Association and Correction.}
For the hypothesis $\hypo{t}{i}$, we select an observation $\obs{t}{i}$ for association as the one that maximizes the following score:
\begin{equation}
\obs{t}{i} = \underset{\obs{t}{} \in \observset{t}{}}{\argmax} \left ( p_{mask} \left ( \obs{t}{} , \hypo{t}{i} \right ) \cdot p_{motion} \left ( \obs{t}{} , \hypo{t}{i} \right ) \right ),
\end{equation}
where $p_{\text{motion}} \left ( \obs{}{i} , \hypo{}{i} \right ) = \mathcal{N} \left( \pose_{obs} \mid \pose_h, \Sigma_h \right )$  is the motion model term,
that evaluates the likelihood of the observation position given the Kalman filter prediction.
The appearance term $p_{mask} \left ( \obs{}{i} , \hypo{}{i} \right ) = \iou \left ( \mask^t_{prio, i} , \obs{t}{} \right )$ scores how well the predicted mask fits to the observation mask
by computing the per-pixel intersection-over-union (IoU) in the image domain.
This way, we jointly utilize 3D space, motion and image information for data association.

Then, the position and velocity $ \left[ \pose^t_{prio}, \velocity^t_{prio} \right] $ is updated via the Kalman filter correction step using the associated observation and the ground-plane estimate.
Since objects are often occluded, the ground-plane estimate is used to obtain the bottom-point by projecting the estimated median position $\pose$ to the ground.
The current mask is replaced by the mask of the new observation, or the predicted mask $ \mask^t_{prio, i} $ in case an observation is missing.

\PAR{Maintaining a Feasible Hypothesis Set.}
With this procedure, we may end up with several duplicates in our hypothesis set,
which we remove using the following method.
We consider a temporal window from $t-\tau$ to $t$ ($\tau$ is in practice set to 10).
We apply non-maximum suppression to the hypotheses which were started before frame $t-\tau$.
As a similarity function we use:
\begin{equation}
	\operatorname{sim} \! \left( \hypo{}{i}, \hypo{}{j} \right) = \frac{\sum_{ t \in T \left( \hypo{}{i}, \hypo{}{j} \right) } \left ( \iou \left ( \mask_i^t, \mask_h^t \right ) \right )}{\left | T \! \left( \hypo{}{i}, \hypo{}{j} \right) \right |},
\end{equation}
where $T \! \left( \hypo{}{i}, \hypo{}{j} \right)$ denotes the set of common time frames of the two hypotheses.
This way, we allow alternative explanations of the data to develop, and only prune them once they had the chance to diverge from the other hypotheses.

\subsection{Inference} At this stage, we obtain several conflicting segmentation hypotheses in the time domain, including over- and under-segmentations, as well as outliers from the static background, such as segments that are parts of buildings.
We perform near-online inference using a CRF model to resolve this ambiguity.
Intuitively, hypotheses that are supported by consistent segmentations are more prominent object candidates.
Two additional cues that we use are the segmentation scores of the associated segmentation masks and the classification scores.
As a counterweight, hypotheses are mutually penalized for occupying the same physical space in the image domain.
We encode these intuitions by performing a MAP inference using a CRF model (\cf \cite{Leibe08TPAMI, Osep17ICRA}) by minimizing the following energy function for each time frame $t_e$:
\begin{equation}
\label{equ:energy_tracking_model}
	\mathbf{E}(\binvec^{t_e}, \hyposet) = \sum_{h_i \in \hyposet} b_i \vartheta \left( h_i, t_e \right) + \sum_{h_i,h_j \in \hyposet} b_i b_j \psi \left( h_i, h_j, t_e \right ),
\end{equation}
where the binary indicator vector $\binvec^{t_e} \in \left \{ 0, 1 \right \}^{\left | \hyposet \right |}$ picks the ''selected'' hypotheses ($ b_i = 1 $ for selected ones).

The unary potential function computes the confidence that a hypothesis $h_i$ represents an actual object,
by summing up contributions from all frames of the hypothesis, and weighting them according to $ e^{ - | t - t_e | / \lambda } $:
\begin{equation}
	\begin{aligned}
		\vartheta \left( h_i, t_e \right)
		= w_{\text{min}}^h
		& - \sum_{t} e^{ - \tfrac{ | t - t_e | }{ \lambda } } \left ( w_{\text{sim}} \Phi_{\text{sim}} \left( \obs{t}{i}, h_i  \right ) + w_{\text{seg}} s_i^t \right ) \\
		& - \sum_{t} e^{ - \tfrac{ | t - t_e | }{ \lambda } } \left ( w_{\text{sem}} \Phi_{\text{sem}} \left( \bbox_2d^t \right ) \right ) .
	\end{aligned}
	\label{equ:unary}
\end{equation}
Here, $\lambda$ is a temporal decay parameter.
The minimal required hypothesis score is given by $w_{\text{min}}^h$ and $w_{\text{sim}}$, $w_{\text{seg}}$, and $w_{\text{sem}}$ are weights for the different terms.
The similarity function $\Phi_{\text{sim}} \left( \obs{t}{i}, h_i  \right )$ between the hypothesis $ h_i $ and the associated observation $ \obs{t}{i} $ at time $t$
is simply a mask IoU between the hypothesis prediction at time $t$ and the observation.
$s_i^t$ represents the score of the segmentation mask.
Given an object classifier, we denote the classifier score with $c_t$.
Then, the semantic term is a truncated classification score:
\begin{equation}
\Phi_{\text{sem}} \left( \bbox_2d^t \right ) = \left\{\begin{matrix}
 c_t, & c_t > C_{min} \\
 0, & \text{else}
\end{matrix}\right.,
\end{equation}
where $C_{min}$ is the minimal classifier score needed for contribution to the total score.
This way, we encourage the selection of hypotheses for which the classifier provides confident responses, while the selection of non-recognized objects is not discouraged.
The pairwise potential function penalizes physical overlap using segmentation masks:
\begin{equation}
\psi \left( {h_i, h_j, t_e} \right ) =  \sum_{t} e^{ - \tfrac{ | t - t_e | }{ \lambda } } \left ( \frac {\left | \mathbf{m}_i^t \cap \mathbf{m}_j^t  \right |} { \text{min} \left ( \left |  \mathbf{m}_i^t \right |,  \left |  \mathbf{m}_j^t \right | \right )} \right ).
\end{equation}
Here, $\left | \cdot \right |$ is the mask area and $\cap$ the mask intersection operator in the image domain.
We do not use IoU as an overlap criterion, which is a common approach.
Instead, we employ intersection over the size of the smaller mask, so that objects of different sizes can still suppress each other.
This way, over-segmentations can be suppressed by the larger object.

The pairwise potentials in (\ref{equ:energy_tracking_model}) are not submodular and hence the problem is NP-hard. However, an approximate solution can be computed efficiently using the multi-branch method by \cite{Schindler06ECCV}.

\subsection{Representation}

Nearly all vision-based multi-object tracking approaches represent the tracked objects using bounding boxes in the image domain \cite{Milan14TPAMI, Yoon16CVPR, Choi15ICCV}.
In the area of image-based single object tracking, there was recently a shift towards mask (pixel-precise) representations \cite{Perazzi16CVPR, Voigtlaender17BMVC}.
Tracking approaches using RGBD, stereo or depth-only data (such as LiDAR) typically represent objects with centroids or 3D bounding boxes \cite{Kaestner12ICRA, Teichman11ICRA, Osep17ICRA}.
More accurate representations include point-clouds \cite{Held14RSS} or fixed-dimensional 3D representations such as GCT \cite{Mitzel12ECCV, Mitzel15ICRA} or voxelgrids \cite{Mitzel12ECCV, Osep17ICRA},
that allow to integrate 3D measurements over time.
However, these representations come at the cost of additional processing time and memory consumption, which are major concerns for multi-object trackers that maintain a large hypothesis space.

We propose to represent the tracked objects by their 3D positions in the world-space and their pixel masks in the image domain.
The masks are stored efficiently with run-length encoding (RLE, which typically reduces the storage size to $\mathcal{O}(\sqrt{n})$, where $n$ corresponds to the number of pixels/points representing the mask).
Most operations can be efficiently performed directly on this compressed representation (\eg intersection-over-union or obtaining a bounding box),
hence there is in general no need to decompress this representation during tracking - except for the mask prediction step.

\subsection{Parameter Training}

We train all hyperparameters of the system in two stages.
Firstly, we optimize the hyperparameters for the hypothesis generation, using the temporal coverage criterion.
Temporal coverage $tc_{\text{GT}}$ for the ground-truth object $\text{GT}$ at time $t$ is defined as:
\begin{equation}
tc_{\text{GT}} = \underset{h}{\text{max}} \left (  \sum_{\tau=t-5}^{t} \left ( e^{ \tfrac{ \tau - t }{ \lambda } } \cdot \iou \left (  \bbox_2d_h^\tau, \bbox_2d_{\text{GT}}^\tau  \right ) \right ) \right )
\end{equation}
here $\iou$ is the intersection-over-union of the bounding boxes, $\bbox_2d_h$ is a hypothesis bounding box, $\bbox_2d_{\text{GT}}$ is the bounding box of the ground-truth object,
and $\lambda$ is a temporal decay parameter.
Intuitively, by maximizing the temporal coverage for all the annotated objects we are preferring parameters that produce hypothesis sets that cover the ground-truth object trajectories within a small temporal window as good as possible.
We regularize this by optimizing the temporal coverage for the $K$ best-scoring hypotheses, within a small temporal window of $6$ time frames.

We optimized the hyperparameters of the potential functions $\vartheta \left( \cdot \right)$ and $\psi \left( \cdot \right )$ used for the inference
for maximal MOTA \cite{Bernardin08JIVP}, by combining the MOTA of the \textit{pedestrian} and \textit{car} categories.
In both cases, we used the Tree of Parzen Estimators method by \cite{Bergstra13ICML}.

\section{Experimental Evaluation}

In this section, we demonstrate that our proposed approach can compete with detection-based baselines in the close camera range on the most frequently appearing object categories.
In addition, we show that the proposed method is able to generalize to other annotated object categories.

We perform the evaluation using the KITTI tracking dataset \cite{Geiger12CVPR}, which is the standard dataset for tracking-by-detection methods in automotive and robotics scenarios.
For classification of the tracked objects, we use the classification component of Faster-RCNN \cite{RenNIPS15} (trained on the COCO dataset).
In order to keep all the components as generic as possible, neither the proposal region generation nor the classification module is fine-tuned on the KITTI dataset. Fine-tuning the proposal generation method and the classifier would lead to better performance, but we are interested in how well our approach can generalize between datasets.

We conducted all experiments using our own splits of the KITTI tracking training set, for which annotations are available \footnote{We used sequences 4, 5, 17, 19 to perform model validation and the rest for the evaluation of our system.}.
This is necessary for two reasons:
a) our proposed approach is stereo-based, and therefore only an evaluation in the close camera range is meaningful and
b) we demonstrate results not only for the \textit{car} and \textit{pedestrian} categories but on other categories as well.
Furthermore, we report the results as a function of the distance from the camera, which the KITTI benchmark does not support.

Note that in the KITTI dataset, only a subset of object categories is annotated: \textit{car, pedestrian, cyclist, van, and truck} (there is an additional \textit{misc} category, containing objects such as car trailers, motorcycles, child strollers, \etc).
Object tracks for categories such as animals, street furniture, bicycles, or street signs would therefore be considered as false positives.

We demonstrate how well our system can track the most common traffic participants with the following experiment: We evaluate tracking performance on the \textit{car} and \textit{pedestrian} categories using the CLEAR MOT metrics \cite{Bernardin08JIVP} on the KITTI dataset. We compare our method to CIWT \cite{Osep17ICRA}, a state-of-the-art tracking-by-detection method using the KITTI evaluation scripts. To evaluate our method on these categories, we report only tracked objects which are recognized as cars or pedestrians. We evaluated CIWT with two different detectors: a) Regionlets \cite{Wang13ICCV} which are the public detections used for the KITTI tracking benchmark, fine-tuned on KITTI (CIWT+Regionlets), and b) Faster-RCNN \cite{RenNIPS15} with SharpMask \cite{PinheiroECCV16} as the proposal generation component (CIWT+SMRCNN).
\begin{figure}
	\includegraphics[width=0.5\textwidth]{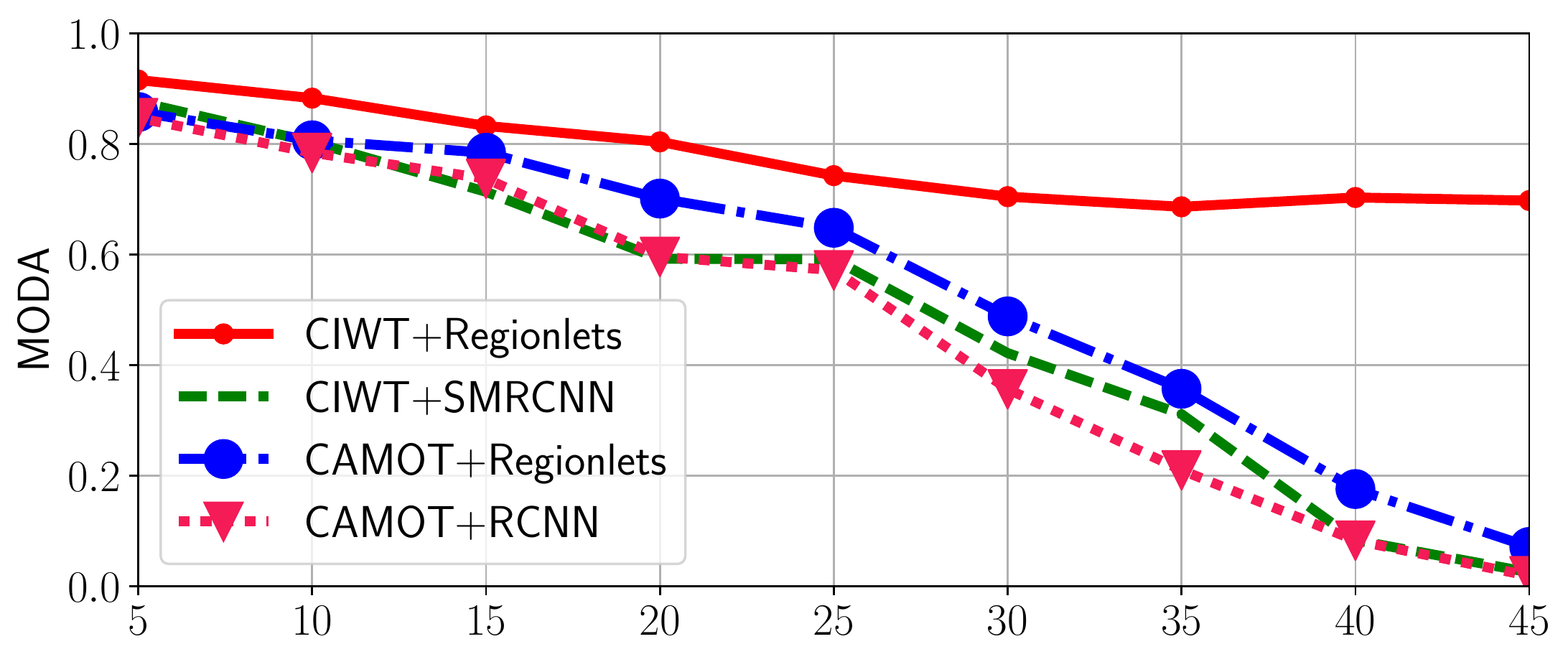}
	\includegraphics[width=0.5\textwidth]{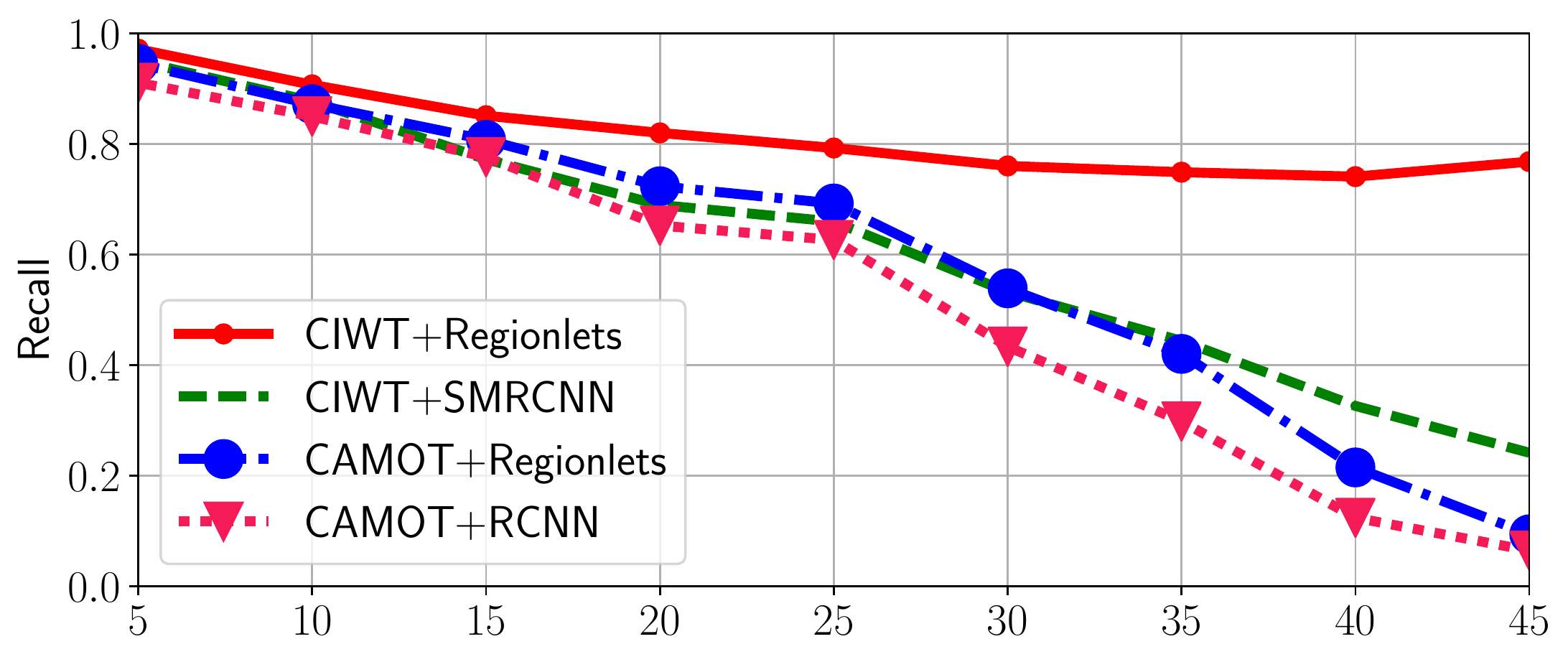}
	\includegraphics[width=0.5\textwidth]{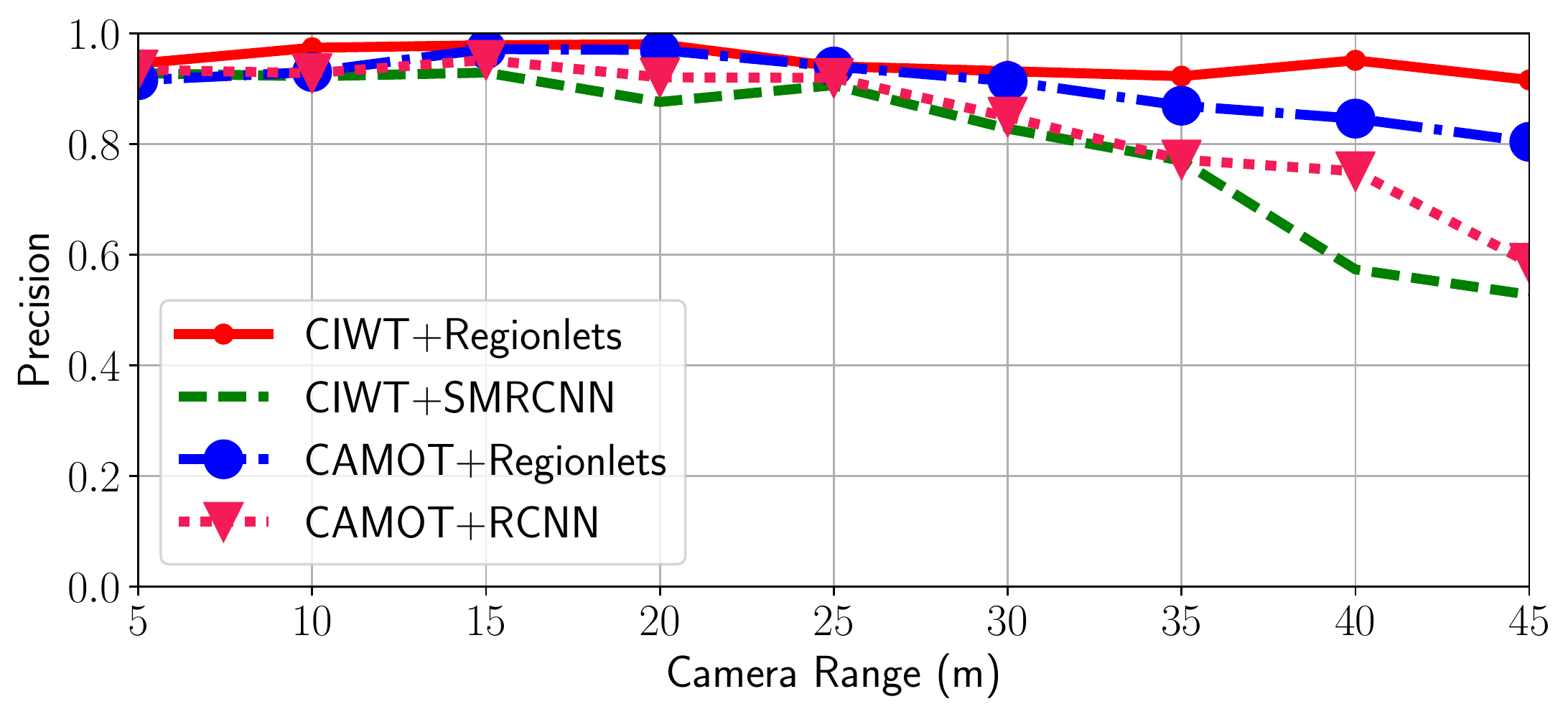}	
	\caption{MODA, Recall and Precision by distance range (higher is better) for cars.}		
		
\label{fig:per_dist_car}
\end{figure}
\begin{figure}
	\includegraphics[width=0.5\textwidth]{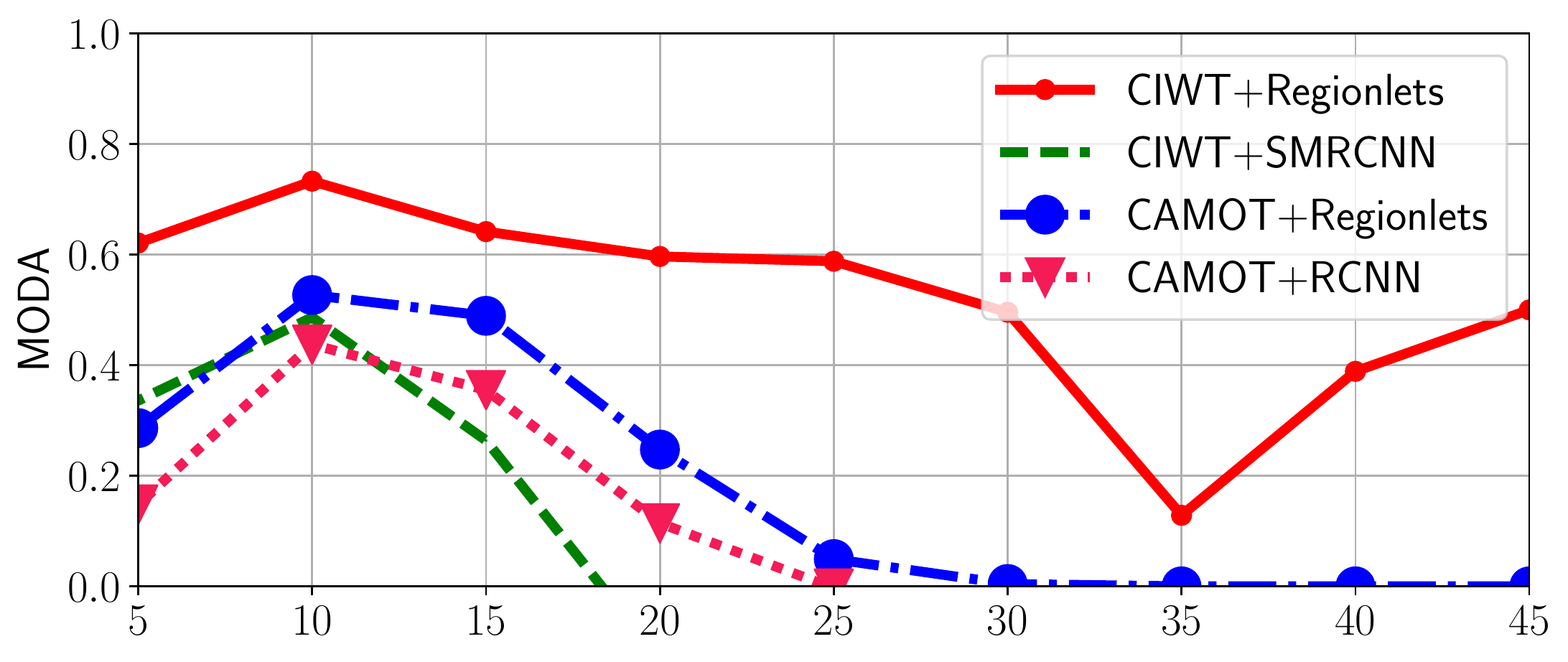}
	\includegraphics[width=0.5\textwidth]{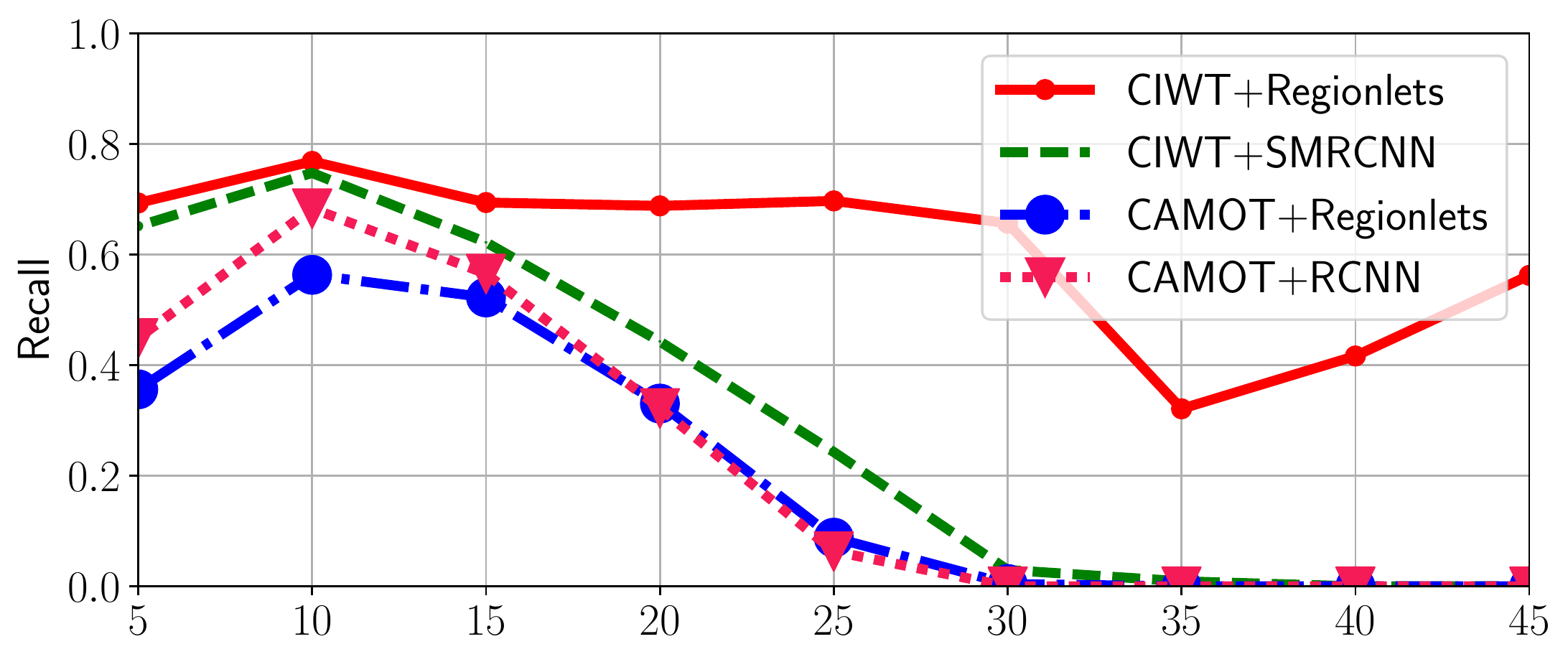}
	\includegraphics[width=0.5\textwidth]{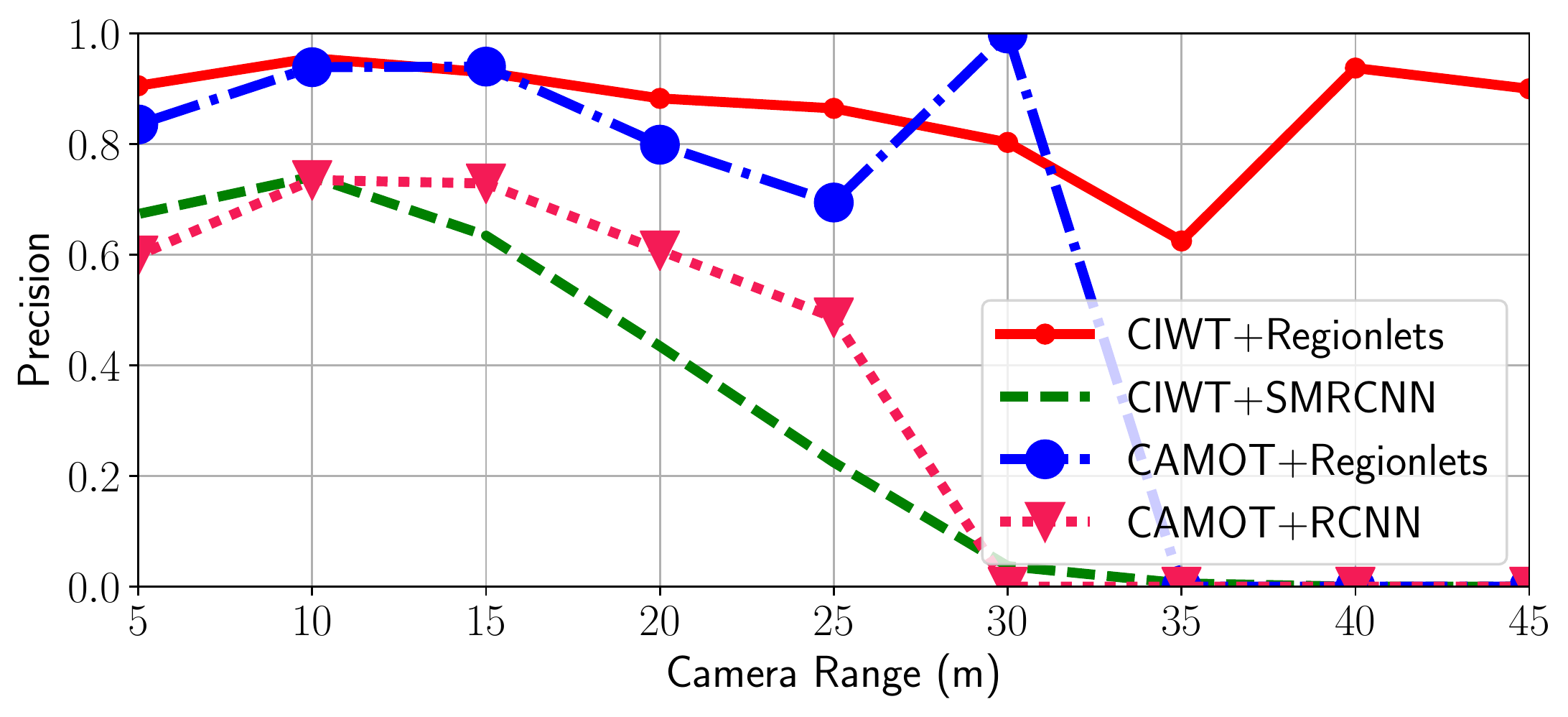}	
	\caption{MODA, Recall and Precision by distance range (higher is better) for pedestrians.}			
\label{fig:per_dist_ped}
\end{figure}

As evident in Figures \ref{fig:per_dist_car} and \ref{fig:per_dist_ped}, the best performer on the \textit{car} and \textit{pedestrian} tracking tasks is the tracking-by-detection method using Regionlets (CIWT+Regionlets), that demonstrates consistently good performance in all camera ranges.
However, both the object detector and the tracker were fine-tuned on KITTI for the specific task of tracking cars and pedestrians.
Perhaps surprising, when using the RCNN-based detector, the tracking performance drops (CIWT+SMRCNN).
We believe that this is due to the fact that these components were not fine-tuned on KITTI.
In general, RCNN is a more powerful detection system than Regionlets.
However, in COCO there is no specific pedestrian category, but only a general person category.
In addition, in COCO cars and persons mainly appear close to the camera, therefore detecting these objects in far distances is challenging.

In the case of our proposed Class-Agnostic Multi-Object Tracker (CAMOT), we used SharpMask (trained on COCO) as the object proposal component for tracking the generic objects (which limits the possibly achievable recall).
For classification of the tracked objects we used a) the classification part of RCNN (CAMOT+RCNN) or b) Regionlet detections (CAMOT+Regionlets).
Here we use the Regionlets as a classifier by matching the detections to the bounding boxes enclosing the masks (in the image domain) and report the IoU between the patches as a classification score.

We obtain better results when using Regionlets to classify our tracked objects (CAMOT+Regionlets) in comparison to RCNN-based verification (CAMOT+RCNN). Again, this is due to the fact that the Regionlets were trained on KITTI. On the other hand, the RCNN classifier is capable of recognizing a much wider range of categories, not only cars, pedestrians, and cyclists.
In addition, CAMOT+Regionlets outperforms the detection-based tracker that uses the non-fine-tuned detector (CIWT+SMRCNN).
We believe that is is the first time any vision-based tracking-before-detection system can compete with a non-fine-tuned tracking-by-detection baseline on specific categories, while still being able to track unknown, arbitrary objects.
(CAMOT+RCNN) is on-par with (CIWT+SMRCNN) up to 20m range, after that, it lags behind.
However, we observed that typically objects are tracked even in farther ranges, but only recognized by the RCNN classifier when they appear closer to the camera.

\begin{figure}
	\includegraphics[width=0.5\textwidth]{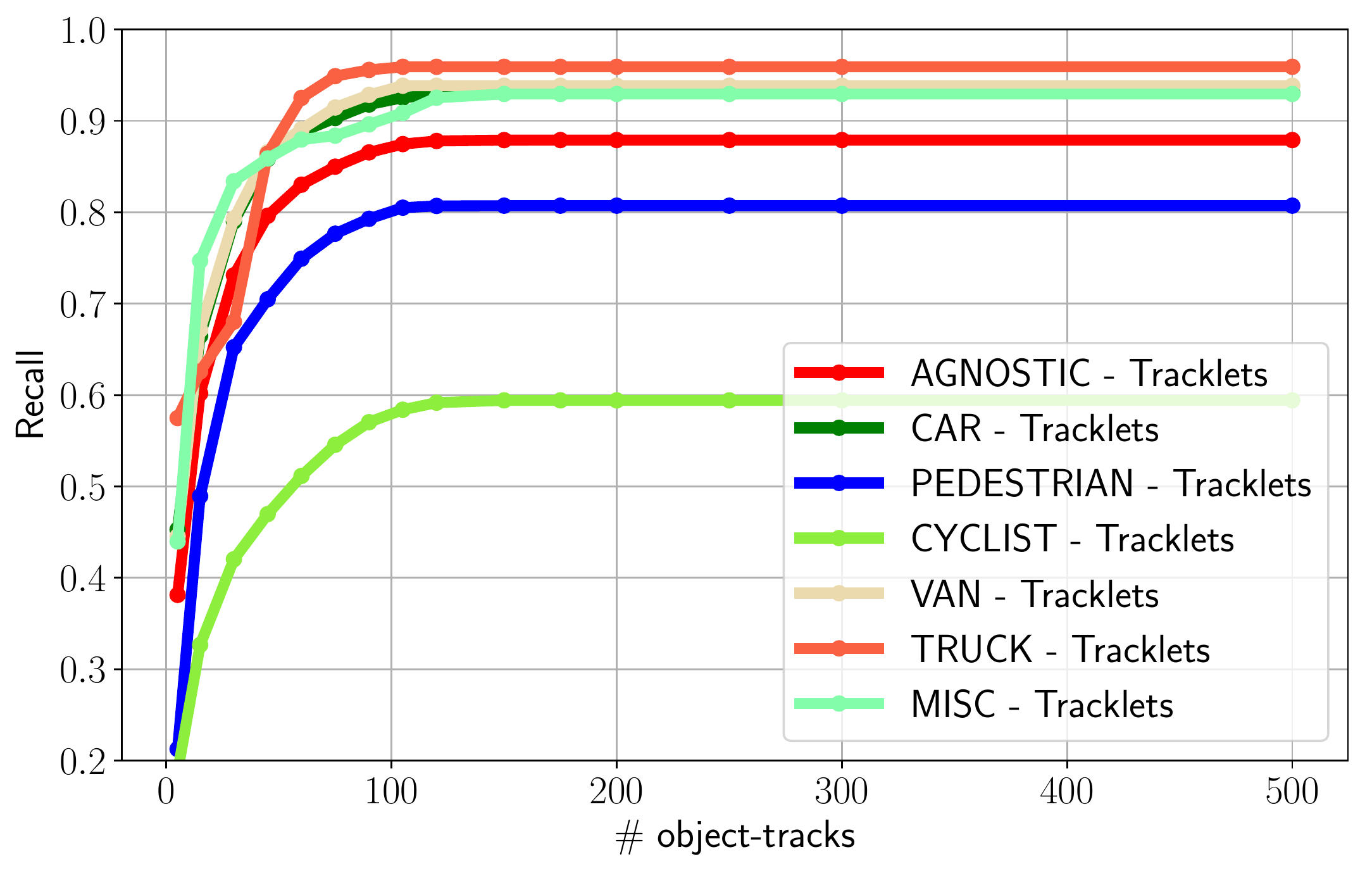}
	\caption{Recall for all categories, annotated in KITTI, covered by our category-agnostic hypotheses.}	
\label{fig:recall_30m}
\end{figure}

\begin{figure}[t!]
\includegraphics[width=0.98\linewidth]{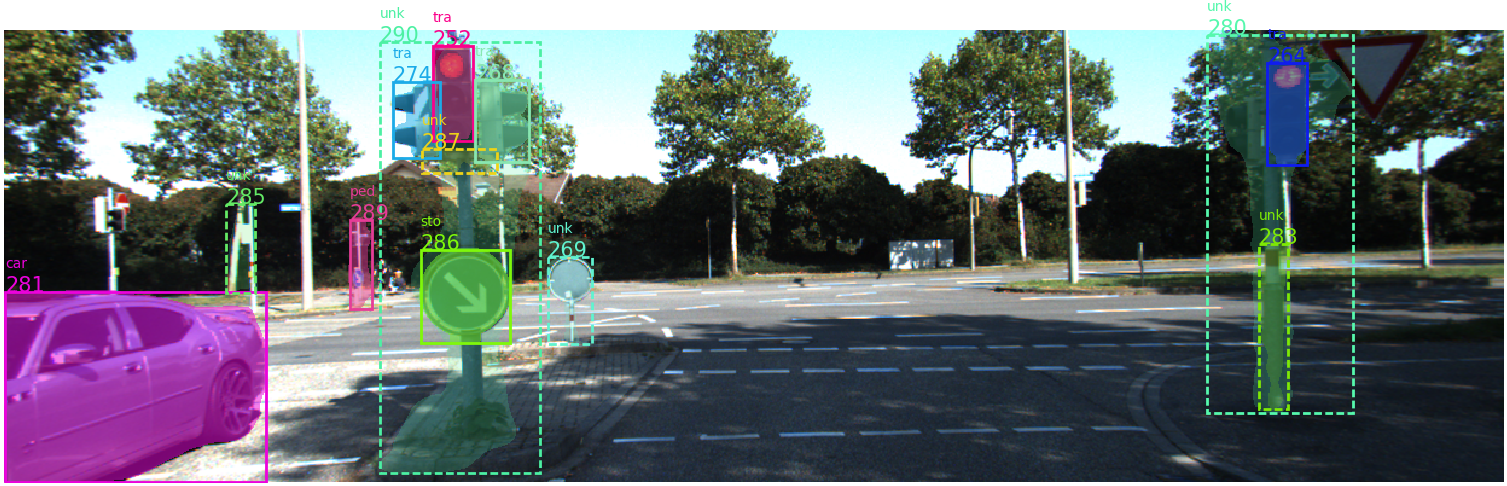}
\includegraphics[width=0.98\linewidth]{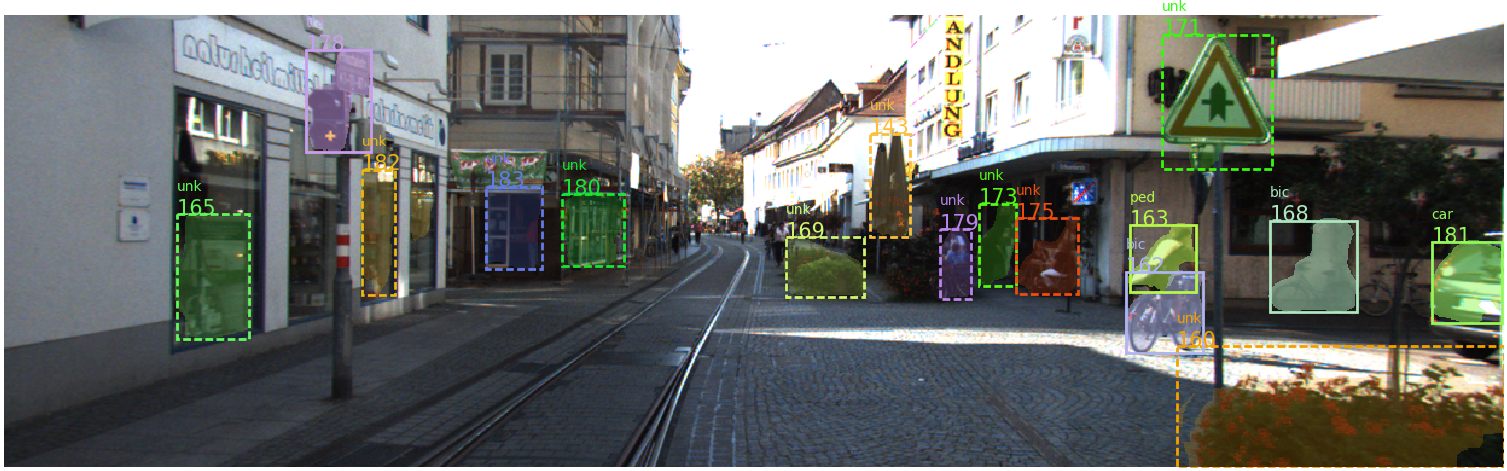}
\includegraphics[width=0.98\linewidth]{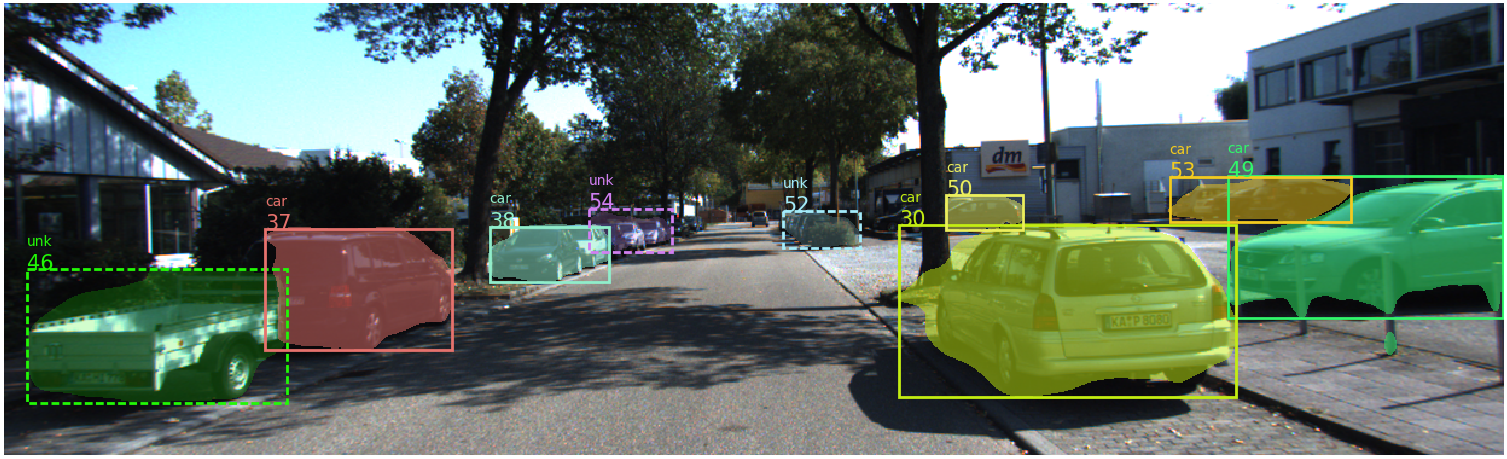}
\includegraphics[width=0.98\linewidth]{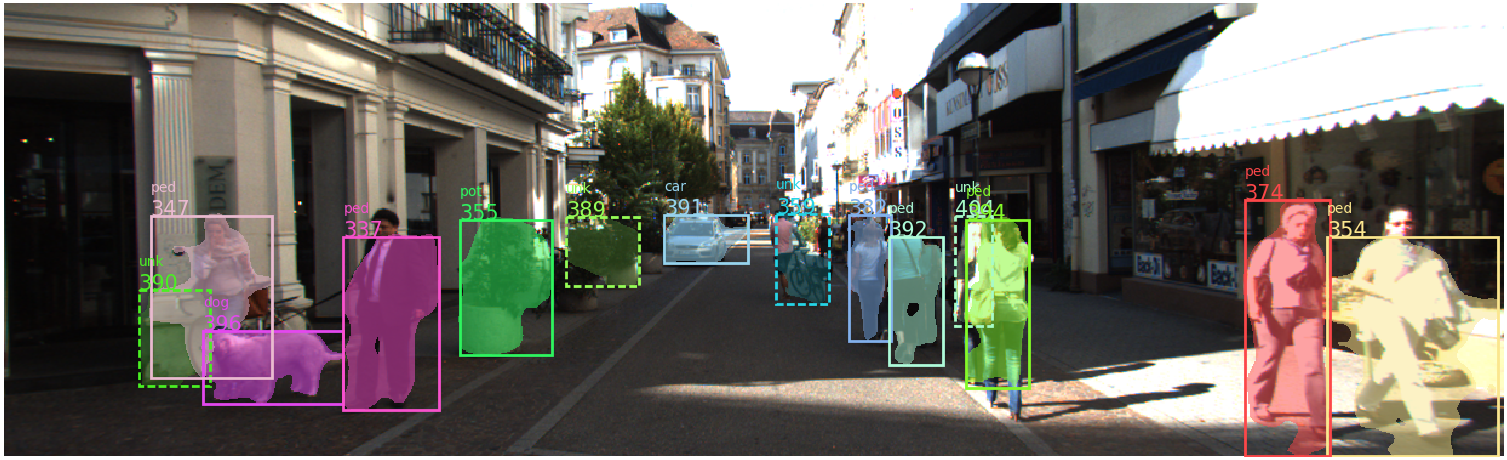}
\includegraphics[width=0.98\linewidth]{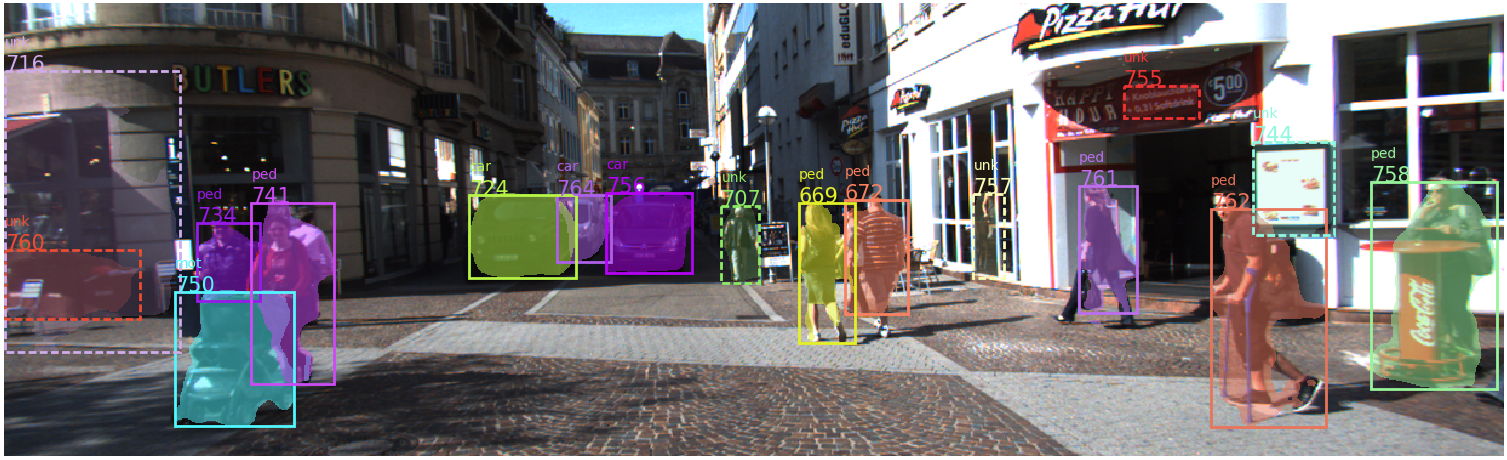}
\includegraphics[width=0.98\linewidth]{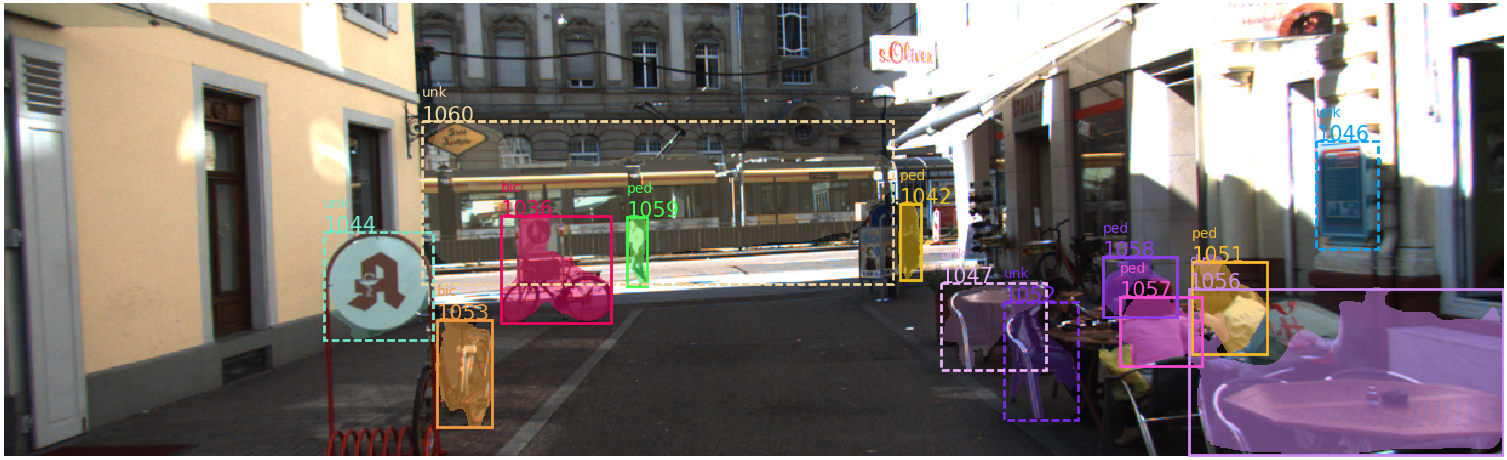}
\includegraphics[width=0.98\linewidth]{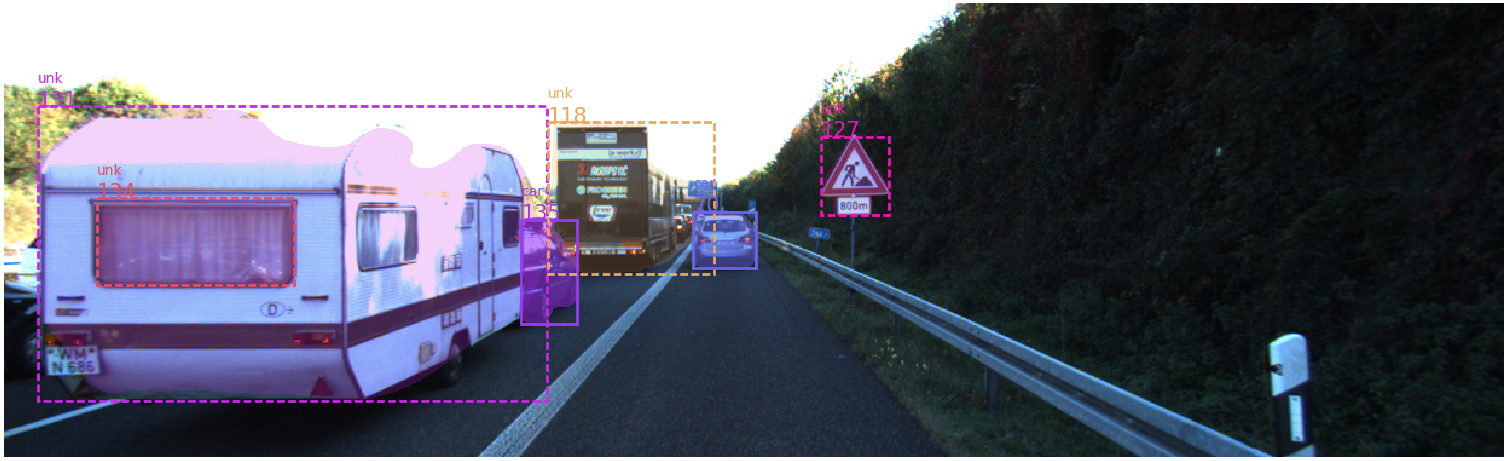}
\caption{Qualitative results on the KITTI tracking dataset. Dashed bounding boxes denote the tracked objects that were not recognized by the classifier.
	We are able to report correct masks for a variety of relevant objects. Besides cars and pedestrians we recognize bikes and other vehicles.
	Traffic signs and lights are also tracked, and in some instances classified.
	Furthermore the dog in the 4th picture and the child stroller in the 5th are reported separately from the surrounding pedestrians.
	We can observe some properties of the method which limit the performance on the benchmark.
	The cyclist and bicycle in the 2nd picture are reported separately.
	In the 3rd picture, some distant cars are merged together. While their spatial extend is estimated correctly, this is undesirable during the evaluation.}
\label{fig:tracking_agnostic}
\end{figure}

Next, we demonstrate that our system is able to generalize to other categories annotated in KITTI.
In Fig. \ref{fig:recall_30m} we evaluate how well all the tracked objects can cover the annotated categories.
We rank the (agnostic) objects tracks using the unary potential function (\ref{equ:unary}) (without the semantic term, \ie setting $w_{sem}=0$).
We achieve very promising results on all the categories.
While it can be argued that for most of these categories good detectors already exist, we point out that we achieve also very good recall on the \textit{misc} category.
The only outlier here is the bicyclist category - we observed that SharpMask tends to produce separate proposals for the bicycle and the rider and we usually track them separately.
We do not consider this to be incorrect. However, in the KITTI annotations the bicycle and the rider are annotated as single bounding box (\textit{cyclist}).
Finally, in Fig. \ref{fig:tracking_agnostic} we show qualitatively that our system is capable of tracking both known and unknown objects simultaneously.
Objects are visualized with their pixel masks and encapsulating bounding boxes.

\section{Conclusion}

In this paper, we presented CAMOT, a novel method for tracking arbitrary objects from a mobile robotics platform. In addition to be able to track the most common traffic participants, we demonstrate that we are able to track arbitrary objects based on category-agnostic proposals. Our system is a step towards a more generic tracking method, that is able to operate in rich environments, in which a robot may encounter previously unseen objects, for which detectors may be difficult to obtain.

In future work, we plan to use our system for object discovery by finding common patterns in the tracked objects, that were not recognized by the classifier. This way, a mobile platform could explore its surroundings and learn about novel objects in an unsupervised or weakly-supervised fashion.

%
\footnotesize {\PAR{Acknowledgments:} This work was funded by ERC
Starting Grant project CV-SUPER (ERC-2012-StG-307432).}

{\small
\bibliographystyle{ieee}
\bibliography{abbrev_short,egbib}

\begin{thebibliography}{10}\itemsep=-1pt

\bibitem{Alexe12TPAMI}
B.~Alexe, T.~Deselaers, and V.~Ferrari.
\newblock Measuring the objectness of image windows.
\newblock {\em PAMI}, 34(11):2189--2202, 2012.

\bibitem{Bergstra13ICML}
J.~Bergstra, D.~Yamins, and D.~D. Cox.
\newblock Making a science of model search: Hyperparameter optimization in
  hundreds of dimensions for vision architectures.
\newblock In {\em ICML}, 2013.

\bibitem{Bernardin08JIVP}
K.~Bernardin and R.~Stiefelhagen.
\newblock Evaluating multiple object tracking performance: The {CLEAR MOT}
  metrics.
\newblock {\em JVIP}, 2008, 2008.

\bibitem{Besl92PAMI}
P.~J. Besl and N.~D. McKay.
\newblock {A method for registration of 3-D shapes}.
\newblock {\em PAMI}, 14(2):239--256, 1992.

\bibitem{Bogoslavskyi16IROS}
I.~Bogoslavskyi and C.~Stachniss.
\newblock Fast range image-based segmentation of sparse 3d laser scans for
  online operation.
\newblock In {\em IROS}, 2016.

\bibitem{Choi15ICCV}
W.~Choi.
\newblock Near-online multi-target tracking with aggregated local flow
  descriptor.
\newblock In {\em ICCV}, 2015.

\bibitem{Dewan15ICRA}
A.~Dewan, T.~Caselitz, G.~D. Tipaldi, and W.~Burgard.
\newblock Motion-based detection and tracking in {3D LIDAR} scans.
\newblock In {\em ICRA}, 2015.

\bibitem{Ester96KDD}
M.~Ester, H.~peter Kriegel, J.~Sander, and X.~Xu.
\newblock A density-based algorithm for discovering clusters in large spatial
  databases with noise.
\newblock In {\em Int. Conf. on Knowledge Discovery and Data Mining}, 1996.

\bibitem{Fischler81ACM}
M.~A. Fischler and R.~C. Bolles.
\newblock Random sample consensus: A paradigm for model fitting with
  applications to image analysis and automated cartography.
\newblock {\em Comm. of the ACM}, 24(6):381--395, 1981.

\bibitem{Geiger12CVPR}
A.~Geiger, P.~Lenz, and R.~Urtasun.
\newblock Are we ready for autonomous driving? {The} {KITTI Vision Benchmark
  Suite}.
\newblock In {\em CVPR}, 2012.

\bibitem{Geiger11IV}
A.~Geiger, J.~Ziegler, and C.~Stiller.
\newblock {StereoScan}: Dense 3d reconstruction in real-time.
\newblock In {\em Intel. Vehicles Symp.'11}, 2011.

\bibitem{Held16RSS}
D.~Held, D.~Guillory, B.~Rebsamen, S.~Thrun, and S.~Savarese.
\newblock A probabilistic framework for real-time 3d segmentation using
  spatial, temporal, and semantic cues.
\newblock In {\em RSS}, 2016.

\bibitem{Held14RSS}
D.~Held, J.~Levinson, S.~Thrun, and S.~Savarese.
\newblock Combining 3d shape, color, and motion for robust anytime tracking.
\newblock In {\em RSS}, 2014.

\bibitem{Horbert2015ICRA}
E.~Horbert, G.~M. García, S.~Frintrop, and B.~Leibe.
\newblock Sequence-level object candidates based on saliency for generic object
  recognition on mobile systems.
\newblock In {\em ICRA}, 2015.

\bibitem{Kaestner12ICRA}
R.~Kaestner, J.~Maye, Y.~Pilat, and R.~Siegwart.
\newblock Generative object detection and tracking in {3D} range data.
\newblock In {\em ICRA}, 2012.

\bibitem{Kang17CVPR}
K.~Kang, H.~Li, T.~Xiao, W.~Ouyang, J.~Yan, X.~Liu, and X.~Wang.
\newblock Object detection in videos with tubelet proposal networks.
\newblock In {\em CVPR}, 2017.

\bibitem{Kochanov2016IROS}
D.~Kochanov, A.~Osep, J.~St\"uckler, and B.~Leibe.
\newblock Scene flow propagation for semantic mapping and object discovery in
  dynamic street scenes.
\newblock In {\em IROS}, 2016.

\bibitem{Leibe08TPAMI}
B.~Leibe, K.~Schindler, N.~Cornelis, and L.~V. Gool.
\newblock Coupled object detection and tracking from static cameras and moving
  vehicles.
\newblock {\em PAMI}, 30(10):1683--1698, 2008.

\bibitem{Lenz2011IV}
P.~Lenz, J.~Ziegler, A.~Geiger, and M.~Roser.
\newblock Sparse scene flow segmentation for moving object detection in urban
  environments.
\newblock In {\em Intel. Vehicles Symp.}, 2011.

\bibitem{Lin14ECCV}
T.-Y. Lin, M.~Maire, S.~Belongie, J.~Hays, P.~Perona, D.~Ramanan,
  P.~Doll{\'a}r, and C.~L. Zitnick.
\newblock Microsoft {COCO}: Common objects in context.
\newblock In {\em ECCV}, 2014.

\bibitem{Liu16ECCV}
W.~Liu, D.~Anguelov, D.~Erhan, C.~Szegedy, S.~Reed, C.-Y. Fu, and A.~Berg.
\newblock {SSD: Single Shot Multibox Detector}.
\newblock In {\em ECCV}, 2016.

\bibitem{Milan14TPAMI}
A.~Milan, S.~Roth, and K.~Schindler.
\newblock Continuous energy minimization for multitarget tracking.
\newblock {\em PAMI}, 36(1):58--72, 2014.

\bibitem{Mitzel15ICRA}
D.~Mitzel, J.~Diesel, A.~O\v{s}ep, U.~Rafi, and B.~Leibe.
\newblock A fixed-dimensional {3D} shape representation for matching partially
  observed objects in street scenes.
\newblock In {\em ICRA}, 2015.

\bibitem{Mitzel12ECCV}
D.~Mitzel and B.~Leibe.
\newblock Taking mobile multi-object tracking to the next level: People,
  unknown objects, and carried items.
\newblock In {\em ECCV}, 2012.

\bibitem{Moosmann13ICRA}
F.~Moosmann and C.~Stiller.
\newblock Joint self-localization and tracking of generic objects in {3D} range
  data.
\newblock In {\em ICRA}, 2013.

\bibitem{Osep16ICRA}
A.~O\v{s}ep, A.~Hermans, F.~Engelmann, D.~Klostermann, M.~Mathias, and
  B.~Leibe.
\newblock Multi-scale object candidates for generic object tracking in street
  scenes.
\newblock In {\em ICRA}, 2016.

\bibitem{Osep17ICRA}
A.~O\v{s}ep, W.~Mehner, M.~Mathias, and B.~Leibe.
\newblock Combined image- and world-space tracking in traffic scenes.
\newblock In {\em ICRA}, 2017.

\bibitem{PinheiroECCV16}
P.~H.~O. Pinheiro, T.~Lin, R.~Collobert, and P.~Doll{\'{a}}r.
\newblock Learning to refine object segments.
\newblock In {\em ECCV}, 2016.

\bibitem{Pinhero16NIPS}
P.~O. Pinheiro, R.~Collobert, and P.~Dollár.
\newblock Learning to segment object candidates.
\newblock In {\em NIPS}, 2015.

\bibitem{Perazzi16CVPR}
J.~Pont-Tuset, F.~Perazzi, S.~Caelles, P.~Arbeláez, A.~{Sorkine}-{Hornung},
  and L.~V. Gool.
\newblock A benchmark dataset and evaluation methodology for video object
  segmentation.
\newblock In {\em CVPR}, 2016.

\bibitem{Redmon16CVPR}
J.~Redmon, S.~Divvala, R.~Girshick, and A.~Farhadi.
\newblock You only look once: Unified, real-time object detection.
\newblock In {\em CVPR}, 2016.

\bibitem{RenNIPS15}
S.~Ren, K.~He, R.~Girshick, and J.~Sun.
\newblock Faster {R-CNN}: Towards real-time object detection with region
  proposal networks.
\newblock In {\em NIPS}, 2015.

\bibitem{Schindler06ECCV}
K.~Schindler, J.~U., and H.~Wang.
\newblock Perspective {N-view} multibody structure-and-motion through model
  selection.
\newblock In {\em ECCV}, 2006.

\bibitem{Teichman11ICRA}
A.~Teichman, J.~Levinson, and S.~Thrun.
\newblock Towards {3D} object recognition via classification of arbitrary
  object tracks.
\newblock In {\em ICRA}, 2011.

\bibitem{Teichman2012IJRR}
A.~Teichman and S.~Thrun.
\newblock Tracking-based semi-supervised learning.
\newblock {\em IJRR}, 31(7):804--818, 2012.

\bibitem{Teichman13ICRA}
A.~Teichman and S.~Thrun.
\newblock Group induction.
\newblock In {\em IROS}, 2013.

\bibitem{Uijlings2013IJCV}
J.~R.~R. Uijlings, K.~E.~A. van~de Sande, T.~Gevers, and A.~W.~M. Smeulders.
\newblock Selective search for object recognition.
\newblock {\em IJCV}, 104(2):154--171, 2013.

\bibitem{Vedaldi08ECCV}
A.~Vedaldi and S.~Soatto.
\newblock Quick shift and kernel methods for mode seeking.
\newblock In {\em ECCV}, 2008.

\bibitem{Vogel13ICCV}
C.~Vogel, K.~Schindler, and S.~Roth.
\newblock Piecewise rigid scene flow.
\newblock In {\em ICCV}, 2013.

\bibitem{Voigtlaender17BMVC}
P.~Voigtlaender and B.~Leibe.
\newblock Online adaptation of convolutional neural networks for video object
  segmentation.
\newblock In {\em BMVC}, 2017.

\bibitem{Wang12ICRA}
D.~Z. Wang, I.~Posner, and P.~Newman.
\newblock What could move? {F}inding cars, pedestrians and bicyclists in {3D}
  laser data.
\newblock In {\em ICRA}, 2012.

\bibitem{Wang15BMVC}
S.~Wang and C.~Fowlkes.
\newblock Learning optimal parameters for multi-target tracking.
\newblock In {\em BMVC}, 2015.

\bibitem{Wang13ICCV}
X.~Wang, M.~Yang, S.~Zhu, and Y.~Lin.
\newblock Regionlets for generic object detection.
\newblock In {\em ICCV}, 2013.

\bibitem{Xiang15ICCV}
Y.~Xiang, A.~Alahi, and S.~Savarese.
\newblock Learning to track: Online multi-object tracking by decision making.
\newblock In {\em ICCV}, 2015.

\bibitem{Yoon16CVPR}
J.~H. Yoon, C.-R. Lee, M.-H. Yang, and K.-J. Yoon.
\newblock Online multi-object tracking via structural constraint event
  aggregation.
\newblock In {\em CVPR}, 2016.

\bibitem{Zhang08CVPR}
L.~Zhang, L.~Yuan, and R.~Nevatia.
\newblock Global data association for multi-object tracking using network
  flows.
\newblock In {\em CVPR}, 2008.

\bibitem{Zhu16ACCV}
G.~Zhu, F.~Porikli, and H.~Li.
\newblock Model-free multiple object tracking with shared proposals.
\newblock In {\em ACCV}, 2016.

\bibitem{Zitnick14ECCV}
C.~L. Zitnick and P.~Doll{\'a}r.
\newblock {Edge Boxes}: Locating object proposals from edges.
\newblock In {\em ECCV}, 2014.

\end{thebibliography}
}

\end{document}